\documentclass[review]{elsarticle}
\usepackage{placeins} 
\usepackage{hyperref}
\usepackage{longtable}
\usepackage{amsmath,amsthm,amssymb,epsfig,bm,amsmath}
 \usepackage{graphicx}
\usepackage{float}
\usepackage{csquotes}
\usepackage{amsmath}
\usepackage{mhchem}
\usepackage{xparse}
\usepackage{adjustbox}
\usepackage{MnSymbol}
\usepackage{mathdots}
\usepackage{makecell}
\usepackage{listings}
\usepackage{footnote}
\usepackage[flushleft]{threeparttable}
\usepackage{multirow}
\usepackage{relsize}
\usepackage{lettrine}
\usepackage{courier}
\usepackage{color} 
\definecolor{mygreen}{RGB}{28,172,0} 
\definecolor{mylilas}{RGB}{170,55,241}
\usepackage{amsmath,amsthm,amssymb,amsfonts} 
\usepackage{graphicx} 
\usepackage[margin=1in]{geometry} 
\usepackage[yyyymmdd]{datetime}
\usepackage{algorithm}
\usepackage[noend]{algpseudocode}
\usepackage{lipsum}
\usepackage{setspace}
\usepackage{siunitx}
\usepackage{pdfpages}
\usepackage{tikz}
\usetikzlibrary{positioning}
\usepackage{booktabs,caption}
\usepackage[]{hyperref}
\usepackage{soul}
\usepackage{xcolor}
\usepackage{subcaption}

\hypersetup{
 colorlinks  = true, 
 urlcolor   = blue, 
 linkcolor  = black, 
 citecolor  = blue 
}
\usepackage[numbers]{natbib}

\usepackage{listings}
\usepackage{color}
\definecolor{codegreen}{rgb}{0,0.6,0}
\definecolor{codegray}{rgb}{0.5,0.5,0.5}
\definecolor{codepurple}{rgb}{0.58,0,0.82}
\definecolor{backcolour}{rgb}{0.95,0.95,0.92}
\lstdefinestyle{mystyle}{
  backgroundcolor=\color{backcolour},  
  commentstyle=\color{codegreen},
  keywordstyle=\color{magenta},
  numberstyle=\tiny\color{codegray},
  stringstyle=\color{codepurple},
  basicstyle=\footnotesize,
  breakatwhitespace=false,     
  breaklines=true,         
  captionpos=b,          
  keepspaces=true,         
  numbers=left,          
  numbersep=5pt,         
  showspaces=false,        
  showstringspaces=false,
  showtabs=false,         
  tabsize=2,
  escapeinside={<@}{@>},
}
 
\usepackage{booktabs} 
\usepackage{siunitx} 
\usepackage{comment}
\usepackage{mathtools} 
\usepackage{gensymb} 
\usepackage{mhchem} 
\usepackage{fixltx2e} 
\usepackage{bbm}

\usepackage{multirow}

\usepackage{fancyhdr} 
\theoremstyle{definition}

\theoremstyle{definition}

\theoremstyle{remark}

\usepackage{soul} 
\soulregister\cite7
\soulregister\ref7
\soulregister\pageref7
\usepackage{bm}

\usepackage{framed} 

\usepackage{multicol} 

\usepackage{nomencl} 
\usepackage{tikz}
\makenomenclature
\usepackage[resetlabels,labeled]{multibib}

\renewcommand*\nompreamble{\begin{multicols}{2}}

\renewcommand*\nompostamble{\end{multicols}}

\usepackage{amssymb}
\usepackage{pifont}
\definecolor{light-gray}{gray}{0.95}

\makeatletter
\AtBeginDocument{\let\hl\@firstofone}
\makeatother

\usepackage[left, pagewise]{lineno}

\journal{Elsevier}

\begin{document}


\begin{frontmatter}

\title{\large On the Impact of Language Nuances on Sentiment Analysis with Large Language Models: Paraphrasing, Sarcasm, and Emojis}

\author{Naman Bhargava$^{a}$, Mohammed I. Radaideh$^{b}$, O Hwang Kwon$^{c}$, Aditi Verma$^{d}$, Majdi I. Radaideh$^{e,*}$}

\cortext[mycorrespondingauthor]{Corresponding Author: Majdi I. Radaideh (radaideh@umich.edu)}

\address{$^{a}$Department of Statistics, University of Michigan, Ann Arbor, MI 48109, United States \\ (Email: \url{namanb@umich.edu})}
\address{$^{b}$Department of Mechanical Engineering, University of Michigan, Ann Arbor, MI 48109, United States \\ (Email: \url{malradai@umich.edu})}
\address{$^{c}$Department of Nuclear Engineering and Radiological Sciences, University of Michigan, Ann Arbor, MI 48109, United States \\ (Email: \url{ohwang@umich.edu})}
\address{$^{d}$Department of Nuclear Engineering and Radiological Sciences, University of Michigan, Ann Arbor, MI 48109, United States \\ (Email: \url{aditive@umich.edu})}
\address{$^{e}$Department of Nuclear Engineering and Radiological Sciences, University of Michigan, Ann Arbor, MI 48109, United States \\ (Email: \url{radaideh@umich.edu})}

\begin{abstract}
\small
\hl{Large Language Models (LLMs) have demonstrated impressive performance across various tasks, including sentiment analysis. However, data quality—particularly when sourced from social media—can significantly impact their accuracy.} This research explores how textual nuances, including emojis and sarcasm, affect sentiment analysis , with a particular focus on improving data quality through text paraphrasing techniques. To address the lack of labeled sarcasm data, the authors created a human-labeled dataset of 5929 tweets that enabled the assessment of LLM in various sarcasm contexts. \hl{The results show that when topic-specific datasets, such as those related to nuclear power, are used to finetune LLMs these models are not able to comprehend accurate sentiment in presence of sarcasm due to less diverse text, requiring external interventions like sarcasm removal to boost model accuracy.} Sarcasm removal led to up to 21\% improvement in sentiment accuracy, as LLMs trained on nuclear power-related content struggled with sarcastic tweets, achieving only 30\% accuracy. In contrast, LLMs trained on general tweet datasets, covering a broader range of topics, showed considerable improvements in predicting sentiment for sarcastic tweets (60\% accuracy), indicating that incorporating general text data can enhance sarcasm detection. The study also utilized adversarial text augmentation, showing that creating synthetic text variants by making minor changes significantly increased model robustness and accuracy for sarcastic tweets (approximately 85\%). Additionally, text paraphrasing of tweets with fragmented language transformed around 40\% of the tweets with low-confidence labels into high-confidence ones, improving LLMs sentiment analysis accuracy by 6\%. Finally, emojis did not significantly affect sentiment analysis for nuclear power-related content, suggesting that emojis may only reinforce, rather than reveal, sentiment in certain contexts.

\end{abstract}


\end{frontmatter}

\setstretch{1.4}


\newpage

\section{Introduction}
\label{sec:intro}

The rise of social media has significantly transformed the nature of written communication, blurring the lines between spoken and written language. As a result, various elements of human speech, such as tone, emotion, and intent, are increasingly embedded in text-based posts, \hl{which were not majorly present in older form of communication like newspapers}. One prominent example is sarcasm—a complex form of expression in which individuals communicate their true feelings by deliberately stating the opposite, often in an exaggerated or ironic manner. This linguistic nuance can greatly influence how a message is interpreted, making it essential to detect sarcasm accurately in order to uncover the speaker’s genuine sentiment \cite{joshi2017automatic}.  \hl{Some studies have shown higher sarcasm usage in United States, compared to other countries like China. \cite{96c3d0c5650548b6b3679f218846818f} . Further, sarcasm interpretation also differs across cultures, with it being seen as witty in western culture and considered rude in Asian countries like China \cite{sarc_cul}.}The importance of sarcasm detection is especially pronounced in sentiment analysis tasks, where misinterpreting sarcasm can lead to incorrect conclusions. This challenge is even more pronounced on social media platforms such as X (formerly Twitter), where posts are often brief, lack sufficient context, and are filled with informal language, emojis, memes, and cultural references that make the detection of sarcasm even more difficult \cite{ashwitha2021sarcasm}.

Another increasingly prominent element of digital communication, particularly on social media, is the use of emojis—small pictorial icons that visually represent emotions, objects, or ideas. Emojis have become a widely adopted tool for enhancing written text, allowing users to express emotions, tone, and intent in ways that plain text alone might not fully capture \cite{shiha2017effects}. By supplementing or even replacing words, emojis can provide additional layers of meaning, helping to clarify sentiment and emotional nuance in a concise and intuitive manner. Their ability to bridge the gap between written and spoken communication makes them especially valuable in sentiment analysis and natural language understanding tasks \cite{gupta2023emoji}. 

\hl{A significant portion of data from platforms like X consists of broken sentences, misspelled words, incorrect grammar usage making its comprehension difficult for LLMs.} In such cases, text paraphrasing becomes a valuable tool for enhancing text clarity and quality \cite{palivela2021optimization}. \hl{In our previous research \cite{kwon2024sentiment}, these three language nuances - sarcasm, emoji and text paraphrasing were identified as the potential source leading to inaccurate sentiment analysis of twitter posts through LLMs. Therefore, in this research we try to improve the performance of models by analyzing and mitigating effect of these language nuances.}  

Multiple machine learning and deep learning models have been used to identify sarcasm in text. Classical machine learning models like Support Vector Machine or Naive Bayes depend on explicit feature generation. In contrast, other deep learning models, like transformers-based networks, can extract features implicitly from the input text \cite{Lou2020,Suhaimin2017}. With the introduction of Large Language Models (LLM) in the last few years and their remarkable performance in text generation and classification tasks, researchers have also employed these models for sarcasm detection. It was observed that pre-trained LLM models, like GPT-3, performed worse than other state-of-the-art traditional natural language processing (NLP) models, such as BERT, for sarcasm detection. However, after fine-tuning, LLM outperformed these models \cite{buaroiu2023capable}. The presence of sarcasm in the text may confound the sentiment analysis models. Therefore, to improve the accuracy of these models, different types of features signifying sarcasm are extracted from the text and given as input to the sentiment analysis models \cite{Bouazizi2015}. Other studies have incorporated sarcasm detection following sentiment analysis, adjusting the predicted sentiment based on sarcasm \cite{alita2019analysis}. However, explicit features provided to sentiment analysis may not always be comprehensive enough to capture sarcasm. Furthermore, in some cases, sarcasm may have a positive polarity; therefore, assuming it is always associated with negative emotion may lead to the omission of sarcasm detection in a large section of the text.

A common preprocessing step in most studies is removing emojis before the NLP task. However, recent studies have replaced the emojis by their textual description or extracted emoji embedding \cite{barbieri2016does}, \cite{eisner2016emoji2vec} to study their effect on various NLP tasks. Due to their high expressive power, emojis improved the model's performance in tasks such as sentiment analysis. Further, it was observed that a more coherent textual description of emoji correlated with the tweet's context gave higher accuracy for sentiment analysis \cite{chen2023understanding}. Emojis can reveal sentiments even more explicitly than some parts of the text, such as the entity name. In general, the sentiment analysis accuracy of classical machine learning and deep learning  models improves by including emoji along with the text \cite{lecompte2017sentiment}\cite{lou2020emoji}. However, the impact of emojis on LLMs and sentiment analysis remains underexplored, particularly in the context of fine-tuning and whether these models consistently benefit from including such symbolic elements.

Another form of inconsistency observed in the data collected from X (formerly Twitter) was incorrect/informal English and broken sentences. A data augmentation approach, e.g., text paraphrasing, could be useful to overcome this issue. Text paraphrasing can be defined as restructuring the text while preserving its meaning and emotion \cite{ho2024investigating}. \hl{Recently, studies have observed that employing text paraphrasing improves model performance over sentiment analysis \cite{gul2025advancing}}. Text paraphrasing is also recognized as a potent solution for eliminating the need for a large amount of human-labeled data for sentiment analysis \cite{ho2024investigating}. One notable research gap is the potential role of text paraphrasing in enhancing the performance of LLMs for sentiment analysis, particularly in data labeling tasks where traditional lexicon-based approaches often fail to capture contextual nuances.

Multiple NLP libraries can be used to automate text labeling and reduce the need for manual data labeling, which can be tedious for large datasets. In a previous study \cite{kwon2024sentiment}, an aggregation of seven open-source libraries was used to assign the final sentiment label- TextBlob \cite{textblob}, Vader \cite{vader}, Stanza \cite{stanza}, Pattern \cite{Pattern}, TweetNLP \cite{camacho-collados-etal-2022-tweetnlp}, Twitter-ROBERTa \cite{camacho-collados-etal-2022-tweetnlp}, and PysentiLM \cite{pysentiment}.  Most of these tools either use lexical rules to compute the sentiment of a sentence or deep learning models like transformers. However, the performance of these models in the case of either sarcastic text or informal/broken sentences is highly doubtful, as displayed by the performance of zero-shot transformers and lexical-based models for sarcasm detection in recent studies \cite{buaroiu2023capable}. Further, while labeling the tweets using these libraries, it was observed here \cite{kwon2024sentiment} that the seven libraries disagreed on the labels of a significant chunk of tweets. The agreement of libraries on tweets was aggregated. In some cases, only 3 or 4 libraries agreed upon the label of tweets, i.e., predicted the same label, making it even more difficult for LLMs to predict accurate sentiments of such tweets without manual intervention. For example, this X post/tweet: ``\textit{Not Negative News: Italy and France pen nuclear deal}'' got three votes to be labeled as negative, three votes to be labeled as neutral, and one vote to be labeled as positive. Therefore, a research gap was identified to analyze the effect of text paraphrasing for tweets that do not have enough agreement from these automated labeling libraries to improve further the ground truth quality and sentiment analysis model performance. 

As the discipline-specific dataset presented in this work is derived from the domain of nuclear power—a controversial energy source—it is valuable to highlight prior studies in this area from both natural language processing (NLP) and machine learning (ML) perspectives. From a technical standpoint, machine learning has been extensively applied within the nuclear industry for a range of tasks, including the use of deep neural networks for predicting nuclear accident progression \cite{radaideh2020neural}, digital twin development for nuclear power plants \cite{kochunas2021digital}, reinforcement learning for optimizing nuclear fuel design \cite{radaideh2021physics}, fault prognosis in nuclear systems \cite{khentout2023fault}, deep Gaussian processes for surrogate modeling of nuclear simulations \cite{radaideh2020surrogate}, radiation shielding analysis \cite{husnain2024machine}, and advanced multiphysics modeling with deep learning \cite{radaideh2019combining}, among others \cite{hu2021data, jinia2024intelligent}. On the NLP side, text-to-image generative models have been used to produce realistic and aesthetically appealing images based on nuclear power-related prompts \cite{joynt2024comparative}. Other studies include automatic sentiment analysis of nuclear power discourse on social media using models such as random forests and long short-term memory networks \cite{xu2022automatic}; investigations of public attitudes toward nuclear power in China through integrated social network analysis \cite{gong2022public}; and analyses of public sentiment in the United States regarding issues such as public trust and spent fuel waste management using aggregate survey data \cite{gupta2019tracking, gupta2020exploring} as well as LLMs \cite{kwon2024sentiment}. Despite these efforts, it is noteworthy that none of the aforementioned studies have addressed the nuanced textual characteristics we aim to explore in this work.

Sentiment analysis on social media data faces significant challenges due to various textual nuances, including sarcasm and emojis. These nuances, along with fragmented language and domain-specific contexts, hinder the accuracy and robustness of LLMs when predicting sentiment. Despite advancements, previous studies have focused solely on improving sarcasm detection through training machine learning classifiers based on annotated sarcasm datasets, often overlooking the benefits of data/topic diversity. Moreover, the need for labeled data, particularly for sarcasm analysis, remains a substantial gap in current research. This study addresses these challenges by exploring multiple strategies to improve sentiment analysis accuracy, including the creation of a human-labeled dataset that connects sarcasm with sentiment analysis, text paraphrasing, and adversarial text augmentation. The study's major contributions can be summarized as follows:
\begin{itemize}
    \item Development of a new \textbf{human-labeled} dataset containing both sarcasm and sentiment labels, facilitating the evaluation of LLMs across different sarcasm scenarios and their influence on sentiment analysis accuracy.
    \item Investigating the effect of text paraphrasing on social media data with fragmented language or short text and how it can improve data quality, boosting the confidence and accuracy of LLM sentiment predictions.
    \item Examining new techniques for mitigating LLM inaccuracies due to the presence of sarcasm, such as text paraphrasing to remove sarcasm, comparing LLM performance on domain-specific versus general datasets for predicting sentiment in sarcastic texts, and employing adversarial text augmentation to enhance model robustness and accuracy for sarcastic content by creating synthetic text variants that reduce the sarcasm impact.
    \item Assessing whether emojis impact LLM accuracy in sentiment analysis for domain-specific datasets, determining whether they primarily reinforce existing sentiment or contribute to revealing it.
\end{itemize}

\hl{In summary, by analyzing the textual nuances described above, we want to highlight the importance of the quality of the training data in fine-tuning LLMs and further improve their performance for sentiment analysis.} Section \ref{sec:data} describes different datasets used in this research work. Section \ref{sec:method} explains the methodology adopted in this study for the LLMs and their fine-tuning approach for the purpose of this study. Section \ref{sec:results} presents the findings of this study along with a discussion of these findings. Section \ref{sec:conc} presents concluding remarks and future work avenues.

\section{Data}
\label{sec:data}

\subsection{General Tweets Dataset}
\label{sec:gtd}

To assess the performance of LLMs on identifying sentiment in presence of sarcasm,  an open-source diverse dataset published on Kaggle \cite{kaggle_gt} was utilized. It contains more than 690,000 tweets on a diverse set of topics. The tweets are categorized into three sentiment labels: positive, negative, and neutral. Preprocessing steps were applied, including the removal of special characters and conversion of all text to lowercase. The dataset is fairly balanced, with around 36\% tweets being Positive, 35\% tweets Negative, and around 29\% being neutral. This dataset will serve as a benchmark to ensure that the conclusions are not limited to a specific topic, as might be the case with more specialized datasets like the nuclear power dataset below.

\subsection{Nuclear Power Dataset}
\label{sec:npd}

Our team has collected a dataset containing 1,200,000 nuclear-related tweets to analyze the general public's viewpoint toward nuclear power, which were introduced and analyzed in \cite{kwon2024sentiment,kwon2024using}. Tweets were scraped from X/Twitter, ranging from the year 2008 to 2023. A variety of keywords, such as ``nuclear power", ``nuclear energy", ``nuclear policy", etc, were employed to ensure that the dataset contains tweets related to nuclear power. Through analysis of this data, it was concluded that nuclear power remains a controversial energy source due to the complex relationship between the political implications of the energy and its application as a carbon-free energy source. The positive sentiments toward nuclear power stemmed from its high power density, reliability regardless of weather conditions, environmental benefits, application versatility, and recent innovations and advancements in both fission and fusion technologies, \hl{similar to the findings in related research papers analyzing such aspects \cite{zhan2021development,price2023thermal,mesquita2025advancements}.} Negative sentiments primarily focused on spent fuel management, high capital costs, and safety concerns, as discussed previously in \hl{\cite{stewart2022capital,price2019advanced,radaideh2018criticality,gu2018history}.} This dataset was then used to analyze bias in large language models in this study \cite{radaideh4949090fairness}.

Multiple standard Python libraries were used to analyze these tweets' sentiments and create ground truth labels for training LLM models \cite{kwon2024sentiment,kwon2024using}. However, a sharp disagreement among these libraries was observed in a large chunk of tweets. Therefore, we aim to explore how to improve this disagreement through text paraphrasing using LLM in this study, which was not explored in our prior work.

\section{Methodology}
\label{sec:method}
\subsection{Large Language Models (LLMs)}

LLMs are huge deep learning models comprising parameters on the order of billions. These models have become popular due to their excellent performance on several tasks, such as text generation and summarization.
One of the early models of LLMS is BERT (Bidirectional Encoder Representations from Transformers). It was trained on bidirectional representation of text. BERT has achieved state-of-the-art results on several NLP tasks such as question-answering, natural language understanding, and inference tasks \cite{kenton2019bert}.  Pretraining of BERT involves two steps: (i) predicting random masked tokens in text using bidirectional context, and (ii) predicting the following sentence after the input sentence, which can help to understand the inter-sentence relationships. This step is particularly useful in inference tasks. 

Recently, several new versions of BERT have been introduced. Some of the notable improvements include: (i) DeBERTa \cite{he2020deberta} which uses two separate vector to represent position and content of the word and ALBERT \cite{lan2019albert}, where no. of parameter are reduced to make training faster. In our study, we have fine-tuned three different variants of BERT models to analyze effect of emoji for sentiment analysis. These pretrained models were fine-tuned for emoticon analysis tasks.

Llama2 \cite{touvron2023llama} is an open-source LLM model released by Meta. It has around 7 Billion parameters. It has an underlying transformer architecture along with several improvements like pre-layer normalization of inputs, utilization of SwiGLU activation function instead of ReLU, and utilization of Rotary Embeddings instead of positional. Apart from that, context length was increased to 4096 from 2048 in Llama1 and Grouped Query Attention was used to reduce memory overhead of remembering generated text. 

In this study, apart from Llama2, we have utilized other open-sourced LLMs: MistralAI \cite{jiang2023mistral7b} and Falcon \cite{almazrouei2023falcon}. These models also contain around 7 billion parameters \hl{and were chosen because they are open-source and can be fine-tuned without the need for model compression techniques like quantization, which affected Llama2 performance notably, as concluded in our previous study \cite{kwon2024sentiment}}. Different evaluation metrics- accuracy, precision,recall and F1-score are used to evaluate the performance of LLMs to analyze overall and individual sentiment-label performance for the models. 

\subsection{Human-labelled Dataset for Sarcasm Analysis}
\label{sec:hld}

To assess how well fine-tuned LLMs can predict sentiment in sarcastic text, it was essential to use a dataset with human-annotated labels, avoiding the use of automated labeling tools, which may introduce bias and inaccurate labels. This manually labeled dataset allows for a more reliable evaluation of sarcasm's effect on sentiment analysis.

A randomly selected subset from the nuclear energy dataset described in Section \ref{sec:npd} was used. This subset contains a balanced mix of sarcastic and non-sarcastic tweets. Each tweet was independently annotated by two human judges, who were instructed to assign two types of labels for each tweet:

\begin{itemize} 
    \item Sarcasm Label: Categorized as either \textbf{Sarcastic} or \textbf{Non-Sarcastic}. To reduce subjectivity, annotators followed a consistent definition of sarcasm: \textit{the use of positive or negative language to express the opposite sentiment}. This is a dictionary definition of sarcasm \cite{sarcdef} and prevents individual bias in interpreting sarcasm from affecting the labeling process.
    
    \item Sentiment Label: Assigned as \textbf{Positive}, \textbf{Negative}, or \textbf{Neutral}. Annotators were specifically instructed to judge sentiment in relation to nuclear power—the core theme of the dataset. To maintain neutrality in the case of news headlines, such tweets were explicitly asked to be labeled as \textbf{Neutral}.
\end{itemize}

Once annotation was completed, tweets with disagreement between the two judges on sentiment were removed. The resulting dataset consists of 5,929 tweets.

\begin{figure}[!h]
  \centering
    \centering
    \includegraphics[width=0.55\textwidth]{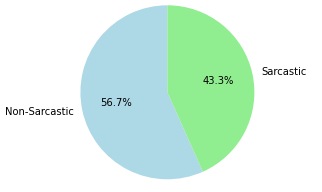} 
    \caption{Distribution of sarcasm labels in the human-labelled dataset} 
    \label{fig:hsa_sarc}
\end{figure}

\begin{figure}[!h]
  \centering
  \begin{subfigure}{0.5\textwidth}
    \centering
    \includegraphics[width=\textwidth]{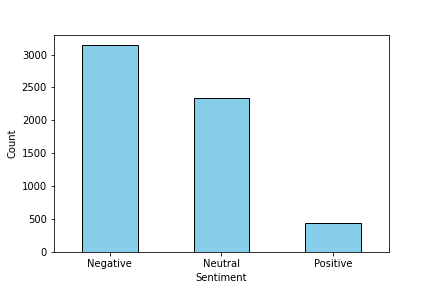}
    \caption{Sentiment distribution for the \textbf{all tweets} in the human-labelled dataset}
    \label{fig:hsa_sent}
  \end{subfigure}
  \hfill
  \begin{subfigure}{0.48\textwidth}
    \centering
    \includegraphics[width=\textwidth]{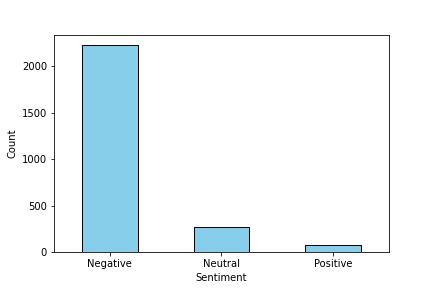}
    \caption{Sentiment distribution for the \textbf{sarcastic} tweets in the human-labelled dataset}
    \label{fig:hsa_sent_sarc}
  \end{subfigure}
  \caption{Sentiment distribution for the human-labeled dataset used for sarcasm analysis}
  \label{fig:human_sarcasm}
\end{figure}

Figure \ref{fig:hsa_sarc} illustrates the distribution of sarcasm labels in the final human-labeled dataset, showing a fairly balanced dataset, with approximately 43\% of tweets classified as sarcastic and 57\% as non-sarcastic. Figure \ref{fig:hsa_sent} presents the sentiment distribution in the human-labeled dataset. The majority of tweets exhibit negative sentiment, followed by neutral and positive sentiments. Figure \ref{fig:hsa_sent_sarc} further breaks down the sentiment distribution for the sarcastic tweets only, revealing that most sarcastic tweets have negative sentiment, with only a few classified as neutral or positive. \textit{Note that the authors prioritized balancing sarcasm in the tweets over sentiment itself in order to determine whether sarcasm can be linked to a specific sentiment in this context.}

\subsection{Sarcasm Effect on Sentiment Analysis}
\label{sec:sarcasm_effect}

This section outlines the methodology used to investigate the impact of sarcasm on the sentiment analysis performance of LLMs, as well as the potential influence of dataset characteristics. Leveraging the previously described human-labeled dataset, we define three distinct fine-tuning scenarios, each with nuanced differences and specific objectives:

\begin{enumerate}
    \item Case 1: LLMs are fine-tuned on high-confidence tweets—specifically, those that received agreement from 5 to 7 of the seven labeling libraries—from the \textbf{Nuclear Power Dataset}. These models are then used to predict sentiment on the human-labeled dataset. Since the human-labeled dataset has tweets with topics similar to the nuclear power dataset, strong performance here would be indicative of effective sentiment generalization and low impact of sarcasm. To evaluate this, the following results are reported for this case:
    \begin{itemize}
        \item Sentiment predictions are made directly on the human-labeled dataset using the fine-tuned LLMs. This is referred to as ``Human Labelled Data Without Sarcasm Augmentation''.
        \item Sarcasm is removed from the tweets in the human-labeled dataset using GPT-3.5 \cite{ye2023comprehensive} by maintaining the same meaning and sentiment without sarcasm, and the same fine-tuned LLMs are then used to predict sentiment. This setup is referred to as ``Human Labeled Data With Sarcasm Augmentation''.
        \item The fine-tuned LLMs are also evaluated solely on the ``sarcastic tweets'' within the human-labeled dataset to determine their effectiveness in handling sarcasm. This is referred to as ``Human Labelled Data Sarcastic Tweets Only''.
    \end{itemize}

    \item Case 2: LLMs are fine-tuned on the \textbf{General Tweet Dataset} described in Section \ref{sec:gtd}, and subsequently used to predict sentiment on the human-labeled dataset. This setup allows us to explore whether training on domain-specific data—such as nuclear power-related content—affects the model’s ability to detect and interpret sarcasm in sentiment analysis. Unlike the nuclear power dataset, the general tweet dataset consists of a wide range of topics and is not domain-specific. As in Case 1, we report the results for the following scenarios: ``Human Labelled Data Without Sarcasm Augmentation'',  ``Human Labelled Data With Sarcasm Augmentation'', and ``Human Labelled Data Sarcastic Tweets Only''.

    \item Case 3: We apply data augmentation to the human-labeled dataset using four easy data augmentation techniques that perturb the original text without changing the original label introduced by \cite{wei2019eda} and can be used by \textbf{TextAttack framework} \cite{morris2020textattack} in Python. This framework leverages augmentation strategies aimed at enhancing NLP models’ robustness against small changes in the text (perturbations) while also promoting better generalization. For each tweet in the human-labeled dataset, five augmented synthetic variants are generated by modifying 10\% of the words in each tweet.

    The original human-labeled data is first split into training and testing subsets with an 80\%/20\% ratio. Only the training portion is processed through the TextAttack framework to produce augmented samples, which are then used to fine-tune the LLMs. The LLMs are subsequently evaluated on the untouched test set from the original human-labeled data.
    
    As in Cases 1 and 2, we report results for the following three scenarios—now based solely on the human-labeled test set (i.e., 20\% of the 5,929 samples):  ``Human-Labeled Data Without Sarcasm Augmentation'',  ``Human-Labeled Data With Sarcasm Augmentation'', and ``Human-Labeled Data Sarcastic Tweets Only''.
\end{enumerate}

The base versions of the LLMs available on Hugging Face are primarily designed for text generation. To adapt them for sentiment classification tasks, we employed Hugging Face’s Sequence Classification API, incorporating an additional linear layer on top for classification purposes. The models were fine-tuned for five epochs using half-precision weights (16-bits) to optimize performance. Training was conducted on four A100 GPUs, courtesy provided by the Idaho National Laboratory computing clusters.

\subsection{Emoji}

The second objective of this study is to investigate the impact of emoticons (or emojis) on sentiment analysis. To begin, we extracted all tweets containing emojis from the nuclear power dataset (refer to Section \ref{sec:npd}) using the Python emoji library \cite{emoji_2025}. This process yielded a total of 77,439 emoji-containing tweets.
Two versions of this subset were created for model training: Version 1 consisted of original tweet with emojis in symbolic form, while in Version 2 emojis which were converted into their textual descriptions using the emoji library. This replacement process involved translating emoji symbols into their corresponding text representations; examples of these mappings are provided in Table \ref{table:emoji_text_mapping}.
Three different variants of BERT models and all the LLMs were fine-tuned on both data versions, and the resulting performance differences are presented in the results section.

\begin{table}[h!]
    \centering
    \caption{Emoji to Text Mapping.}
    \begin{tabular}{|c|c|}
        \hline
        \textbf{Emoji} & \textbf{Text} \\
        \hline
        \includegraphics[width=1cm, height=1cm]{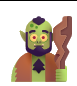} & :troll: \\
        \hline
        \includegraphics[width=1cm, height=1cm]{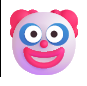} & :clown\_face: \\
        \hline
        \includegraphics[width=1cm, height=1cm]{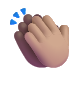} & :clapping\_hands\_medium\_skin\_tone: \\
        \hline
        \includegraphics[width=1cm, height=1cm]{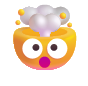} & :exploding\_head: \\
        \hline
    \end{tabular}
    
    \label{table:emoji_text_mapping}
\end{table}

\subsection{Text Paraphrasing}
\label{sec:text_parap}
On examination of data collected from X/Twitter for the nuclear power dataset (see Section \ref{sec:npd}), it was observed that a large number of X posts/tweets contained broken English and incomplete words. This also led to disagreement in their labeling through the seven open-source labeling libraries as described before in Section \ref{sec:intro}. Therefore, we decided to paraphrase the tweets using GPT-3.5 \cite{ye2023comprehensive}, courtesy of the University of Michigan (UM-GPT tools) to ensure better input text quality for LLMs. Based on agreement among the seven labeling libraries (TextBlob, Vader, Stanza, Pattern, TweetNLP, TwitROBERT, and PysentiLM), the tweets were divided into two categories: 
\begin{itemize}
    \item High-Confidence Tweets: \textbf{Five} or more libraries agreed upon the label of these tweets. 
    \item Low-Confidence Tweets: Only \textbf{three} or \textbf{four} libraries agreed upon the label of these tweets.
\end{itemize}

Only X posts/tweets from the \textbf{Low-Confidence} category were paraphrased using GPT-3.5. The following prompt was used to paraphrase the tweets: \textit{``Paraphrase the following text while keeping the response length approximately the same as the original text.''} Keeping the response text length the same helps in optimizing API cost and reduces the possibility of LLM hallucination. 

\section{Results}
\label{sec:results}

\subsection{Text paraphrasing results}
\label{sec:data analysis}

Table \ref{tab:data_comparison} displays a performance comparison of the 7 billion parameter LLMs trained over paraphrased and non-paraphrased datasets based on nuclear power tweets using classification accuracy as an evaluation metric. An improvement in the accuracy of models was observed on augmenting tweets by paraphrasing, with the accuracy of models increasing by 3-6\%. Apart from accuracy, other metrics used to evaluate models are precision, recall, and F1-score. The precision metric indicates how often the model predicts the correct sentiment labels. Recall indicates how often the model can identify the true sentiment label of a tweet among all tweets of a particular sentiment in a dataset. The F1 score is the harmonic mean of both precision and recall and indicates whether the model can perform optimally on both these metrics. Fine-tuned LLMs on paraphrased data have displayed optimal performance on precision, recall, and F1-score evaluation metrics, as shown in Table \ref{tab:model_metrics}. A high precision metric indicates the ability of these models to make a few false positive predictions for each sentiment label, and a high recall metric indicates the ability of these models to identify the true label of most of the tweets. 

\begin{table}[h!]
\centering
\caption{Comparison of classification accuracy between non-paraphrased (original) and paraphrased nuclear power tweets.}
\begin{tabular}{lcc}
\toprule
Model & Non-paraphrased Data (\%) & Paraphrased Data (\%) \\
\midrule
Falcon & 84.0 & 87.1 \\
Mistral & 82.0 & 88.0 \\
Llama-2 & 85.0 & 88.1 \\
\bottomrule
\end{tabular}
\label{tab:data_comparison}
\end{table}

\begin{table}[h!]
\centering
\caption{Performance metrics of LLM models on paraphrased nuclear power tweets.}
\begin{tabular}{lccc}
\toprule
\textbf{Model} & \textbf{Precision (\%)} & \textbf{Recall (\%)} & \textbf{F1-score (\%)} \\
\midrule
Mistral & 86.7& 86.3 & 86.3 \\
Llama-2 & 86.7 & 86.3 & 86.3 \\
Falcon & 85.7 & 86.0 & 85.7 \\
\bottomrule
\end{tabular}
\label{tab:model_metrics}
\end{table}

\begin{figure}[h!]
  \centering
    \centering
    \includegraphics[width=0.6\textwidth]{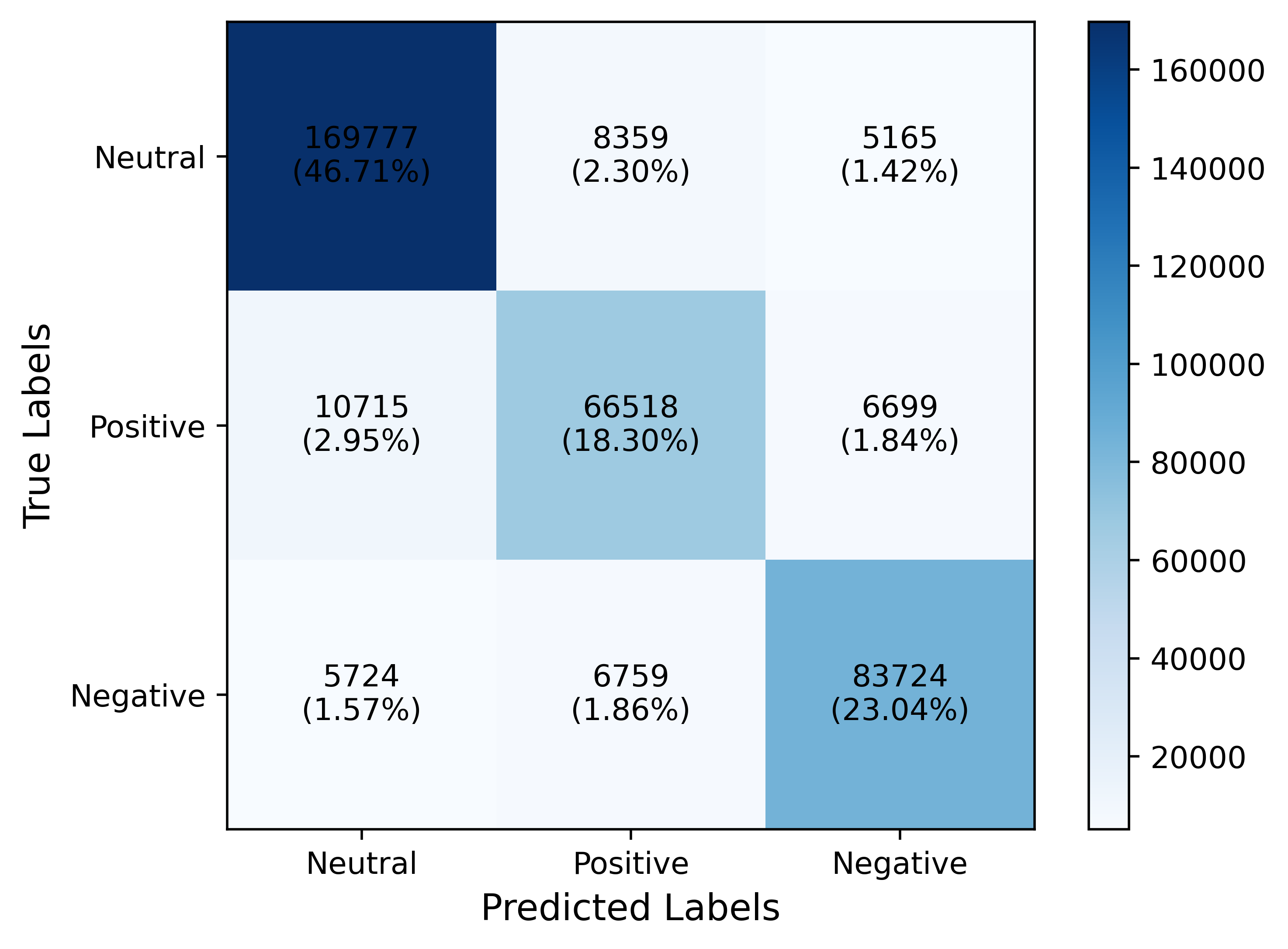} 
    \caption{Llama-2 confusion matrix when fine-tuned on the paraphrased nuclear power tweets.} 
    \label{fig:conf_llm}
\end{figure}

Figure \ref{fig:conf_llm} shows the confusion matrix for the Llama-2 model as a selected LLM with top performance. The x-axis contains true labels, while the y-axis has the predicted labels. We aimed to maximize the diagonal elements: True Positive, True Negative, and True Neutral labels. The percentages are calculated as the number of instances shown in every square over the total number of X posts/tweets (363,440). Llama-2's largest confusion is between positive and neutral sentiments; in 10,715 tweets, Llama-2 confuses the neutral with the positive sentiment and vice versa in 8,359 tweets. This can be due to the significantly lower number of positive tweets used to fine-tune Llama-2 than the neutral tweets, as indicated by our original study \cite{kwon2024sentiment}, where neutral tweets are 51.5\% of all 1.2 million tweets. In contrast, the positive tweets are 15.29\%. The negative tweets are 33.21\% of the total, and the confusion between the negative and positive sentiments is lower than that of neutral. This is related to the label imbalance, where more positive examples are needed to balance the neutral and negative tweets and reduce this confusion. It should be noted that paraphrasing was not meant to change the labels, as will be shown shortly, so the portions of the three sentiments are conserved.  

\begin{figure}[t!]
  \centering
    \centering
    \includegraphics[width=0.55\textwidth]{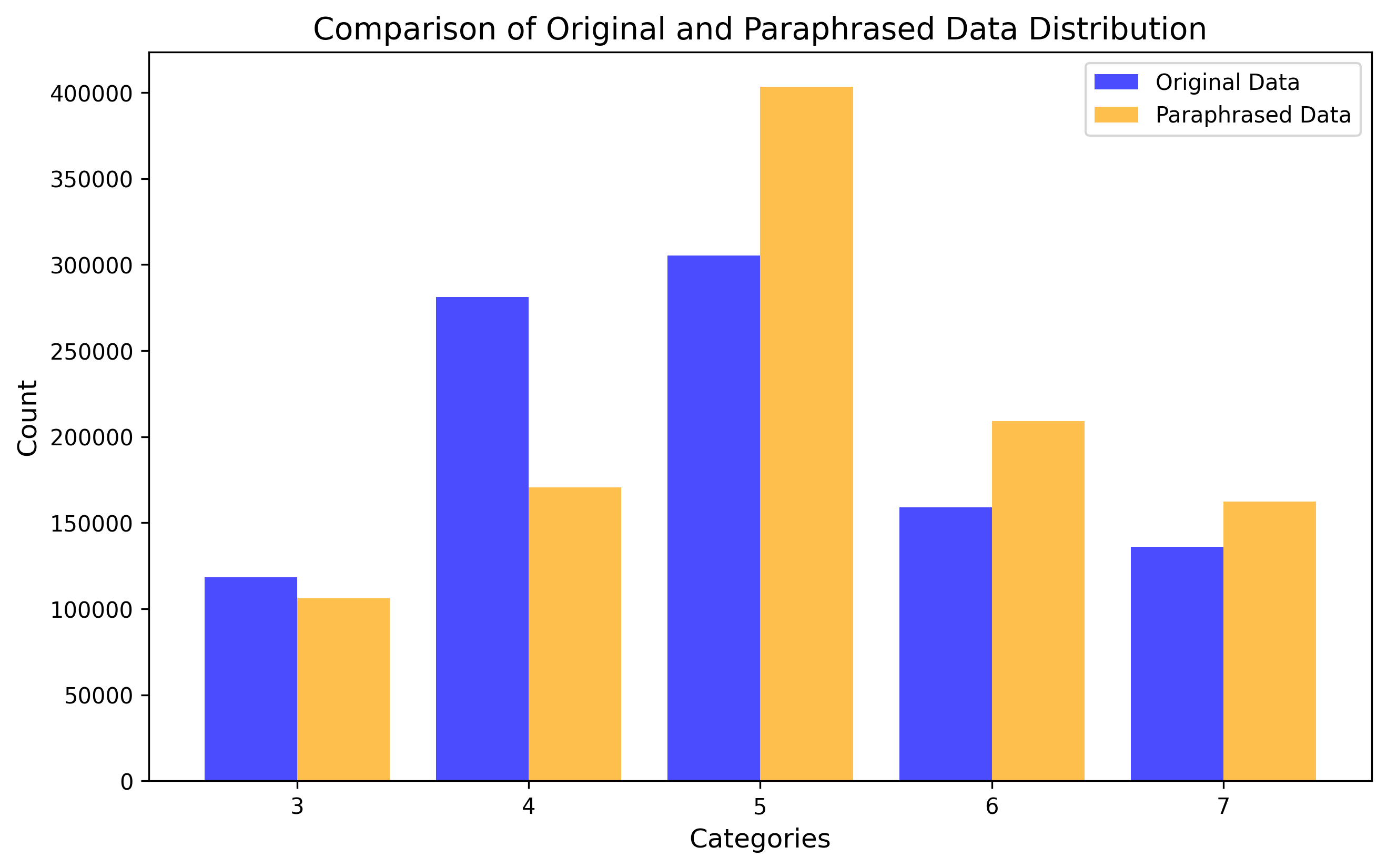} 
    \caption{Library Agreement distribution of original vs paraphrased tweets.} 
    \label{fig:paraphrase_cat}
\end{figure}

Figure \ref{fig:paraphrase_cat} illustrates the level of agreement among sentiment analysis libraries for paraphrased tweets related to nuclear power. Initially, tweets with low confidence were primarily classified under the 3 or 4 library agreement categories as described in Section \ref{sec:text_parap}. However, after paraphrasing, approximately 40\% of the low-confidence tweets have been improved and shifted toward being classified as high-confidence tweets, showing agreement among 5 to 7 libraries. This increased consensus among the seven sentiment labeling libraries, as depicted in Figure \ref{fig:paraphrase_cat}, suggests enhanced data quality. The improved agreement contributes to the better classification accuracy of LLMs, as shown in Table \ref{tab:data_comparison}. Notably, tweets with agreement among 4 libraries showed the greatest decrease in number, while those with 5 library agreements experienced the most significant increase. These findings highlight how poor data quality—such as unclear language, broken English, or incomplete sentences—can negatively impact the performance and fine-tuning of LLMs for sentiment analysis purposes.  

To determine whether GPT-3.5 altered the meaning of the text during paraphrasing, Llama-3 \cite{grattafiori2024llama3} was used as an independent evaluator, separate from both the paraphrasing model (GPT-3.5) and the fine-tuned LLMs. Each original tweet and its paraphrased version were input into Llama-3, which was asked to assess whether the two conveyed the same meaning. Llama-3 found that more than 98\% of the paraphrased tweets preserved the original meaning, indicating that GPT-3.5 paraphrased the text with high semantic accuracy.

\begin{table}[b!]
\centering
\footnotesize
\caption{Original and paraphrased tweet sample comparison from the nuclear power tweet dataset.}
\begin{tabular}{p{4cm} p{4cm} p{5cm}}
\toprule
\textbf{Original Tweet} & \textbf{Paraphrased Tweet} & \textbf{Notes} \\
\midrule
Trump Promising Arms Race Could Set World on Uncertain Path  New York Times  & Trump's pledge to engage in an arms race could lead the world down an unpredictable path, warns the New York Times. & Additional word (\textit{warns}) added which aids in sentiment analysis \\
\midrule
Intercept Perez moved in October to purge longstanding party officials seen as friendly to Sanders and Rep Keith Ellison DMinn while appointing a number of corporate lobbyists including registered lobbyists for Citigroup a nuclear power company      & In October, Perez took action to remove party officials who were perceived as supportive of Sanders and Rep Keith Ellison DMinn, and instead appointed several corporate lobbyists, including those representing Citigroup and a nuclear power company. & Made sentence more coherent \\
\midrule
John Kerry admits Cold Warera nuclear drills in school still conditions my thinking   nato        & John Kerry acknowledges that the nuclear drills conducted during the Cold War era in schools continue to influence his mindset regarding NATO. & Sentence structure improved along with the addition of pronoun to make it more meaningful \\
\midrule
UK doesnt get hydro power from Sweden other than nominally We get gas from Norway but most of the electricity we import comes from France Nucleargenerated paradoxicallyUndersea cables are few in number Seamounted overhead cables are nonexistent     & The UK does not receive significant hydro power from Sweden. Instead, most of our imported electricity comes from France, which is mainly generated by nuclear power. It is worth noting that there are only a few undersea cables and no overhead cables for this purpose. & Multiple sentences in tweet tied together to be more consistent and meaningful \\
\bottomrule
\end{tabular}
\label{tab:text_comparison}
\end{table}

To further understand the effect of paraphrasing, the original and paraphrased tweets were manually inspected and compared in Table \ref{tab:text_comparison}. It was observed that original tweets were written informally and concisely, which might be due to the nature of the usage of X/Twitter by the general public and the word limit enforced with each tweet. The paraphrased version improved the text's grammar by adding punctuation and pronouns. They have also paraphrased the text structure to make it more formal and coherent without adding new information. This was especially observed in the case of tweets with multiple sentences, where paraphrased tweets could make the relation between the sentences more clear and consistent. Further, in a few cases, the paraphrased version also added words that made the tweet's sentiment explicit and easily identifiable. These could be the reasons for improving the performance of LLMs when trained on the paraphrased datasets. Table \ref{tab:text_comparison} shows a few such improvements.

\subsection{Sarcasm Analysis with Fine-tuned LLMs}

Table \ref{tab:sarcasm_np} and Table \ref{tab:sarcasm_gt} compare the performance of LLMs fine-tuned on Nuclear Power (Case 1) and General tweet (Case 2) datasets as described in Section \ref{sec:sarcasm_effect}, respectively. Several insights can be drawn from Tables \ref{tab:sarcasm_np}-\ref{tab:sarcasm_gt}.

\begin{table}[!b]
\centering
\caption{LLM accuracy metrics when trained on the \textbf{Nuclear Power Dataset}. Model is evaluated on Human Labelled Data (row 1), Human Labelled Data augmented by GPT-based text to reduce sarcasm (row 2), and Sarcastic Tweets only from Human Labelled Data (row 3). }
\begin{tabular}{lccc}
\toprule
\textbf{Evaluation Data}& \textbf{Falcon (\%)} & \textbf{Llama-2 (\%)} & \textbf{Mistral (\%)} \\
\midrule
Human Labelled Data Without Sarcasm Augmentation& 50.2& 52.5& 51.9\\
Human Labelled Data With Sarcasm Augmentation& 67.5&   67.5& 69.5\\
Human Labelled Data Sarcastic Tweets Only& 31.3& 36.9& 35.2\\
\bottomrule
\end{tabular}
\label{tab:sarcasm_np}
\end{table}

\begin{table}[!b]
\centering
\caption{LLM accuracy metrics when trained on the \textbf{General Tweets Dataset}. Model is evaluated on Human Labelled Data (row 1), Human Labelled Data augmented by GPT-based text to reduce sarcasm (row 2), and Sarcastic Tweets only from Human Labelled Data (row 3).}
\begin{tabular}{lccc}
\toprule
\textbf{Evaluation Data}& \textbf{Falcon (\%)} & \textbf{Llama-2 (\%)} & \textbf{Mistral (\%)} \\
\midrule
Human Labelled Data Without Sarcasm Augmentation& 43.9& 48.0& 41.3\\
Human Labelled Data With Sarcasm Augmentation& 50.7&   52.6& 47.3\\
Human Labelled Data Sarcastic Tweets Only& 61.5& 73.6& 60.2\\
\bottomrule
\end{tabular}
\label{tab:sarcasm_gt}
\end{table}

Initially, the LLMs demonstrate relatively low accuracy in predicting sentiment on the human-labeled dataset without sarcasm augmentation—achieving around 50\%-52\% accuracy when fine-tuned on nuclear power data, and 41\%-48\% when trained on the general tweet dataset (see the first row of Tables \ref{tab:sarcasm_np} and \ref{tab:sarcasm_gt}).

Second, interestingly, models fine-tuned on the general tweet dataset achieve significantly higher accuracy on sarcastic tweets—up to 74\%—compared to just 30\% accuracy for models trained on nuclear power tweets, as shown in the third row of Tables \ref{tab:sarcasm_np} and \ref{tab:sarcasm_gt}. This suggests that the general tweet dataset provides greater robustness in identifying and handling sarcasm, despite the human-labeled dataset being derived from nuclear power-related content.

Third, and even more notably, the robustness of the general tweet dataset to sarcasm becomes clearer when examining row 2 of Tables \ref{tab:sarcasm_np} and \ref{tab:sarcasm_gt}. Sarcasm augmentation leads to a substantial performance boost for models fine-tuned on the nuclear power dataset, while the improvement is less pronounced for models trained on the general tweet dataset. For example, Falcon's accuracy increased from 50\% to 68\% on the nuclear dataset (compared to a smaller rise from 44\% to 51\% on the general dataset). LLaMA's performance improved from 53\% to 68\% for nuclear tweets (versus 48\% to 53\% for general tweets). Mistral showed a notable increase from 52\% to 70\% on nuclear tweets after sarcasm augmentation but showed comparatively less improvement on general tweets, with accuracy changing from 41\% to 47\%.

\begin{figure}[!h]
  \centering
  \begin{subfigure}{0.7\textwidth}
    \centering
    \includegraphics[width=\textwidth]{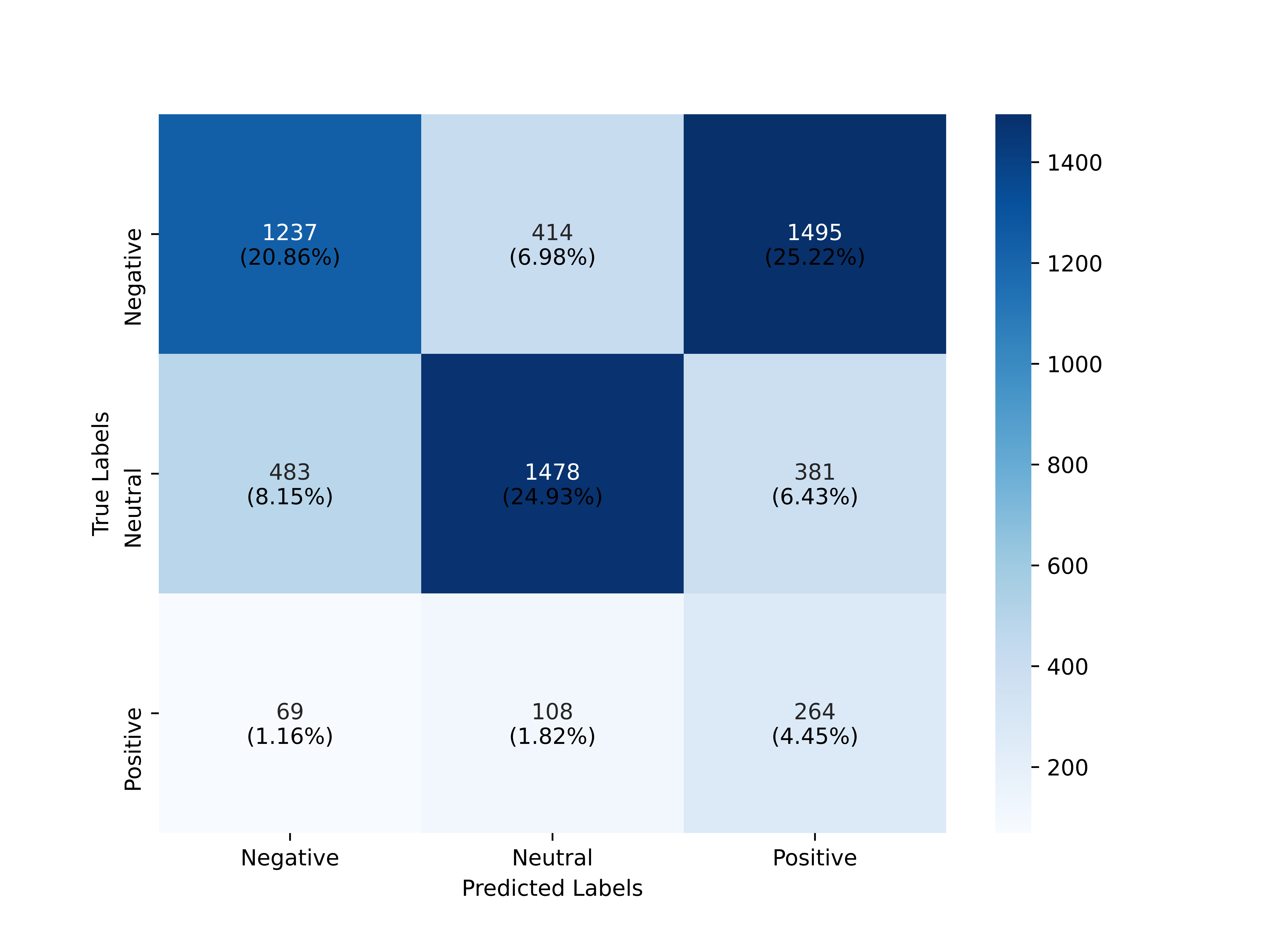}
    \caption{Falcon model evaluated on Human Labelled Data.}
    \label{fig:falcon_hsa}
  \end{subfigure}
  \hfill
  \begin{subfigure}{0.7\textwidth}
    \centering
    \includegraphics[width=\textwidth]{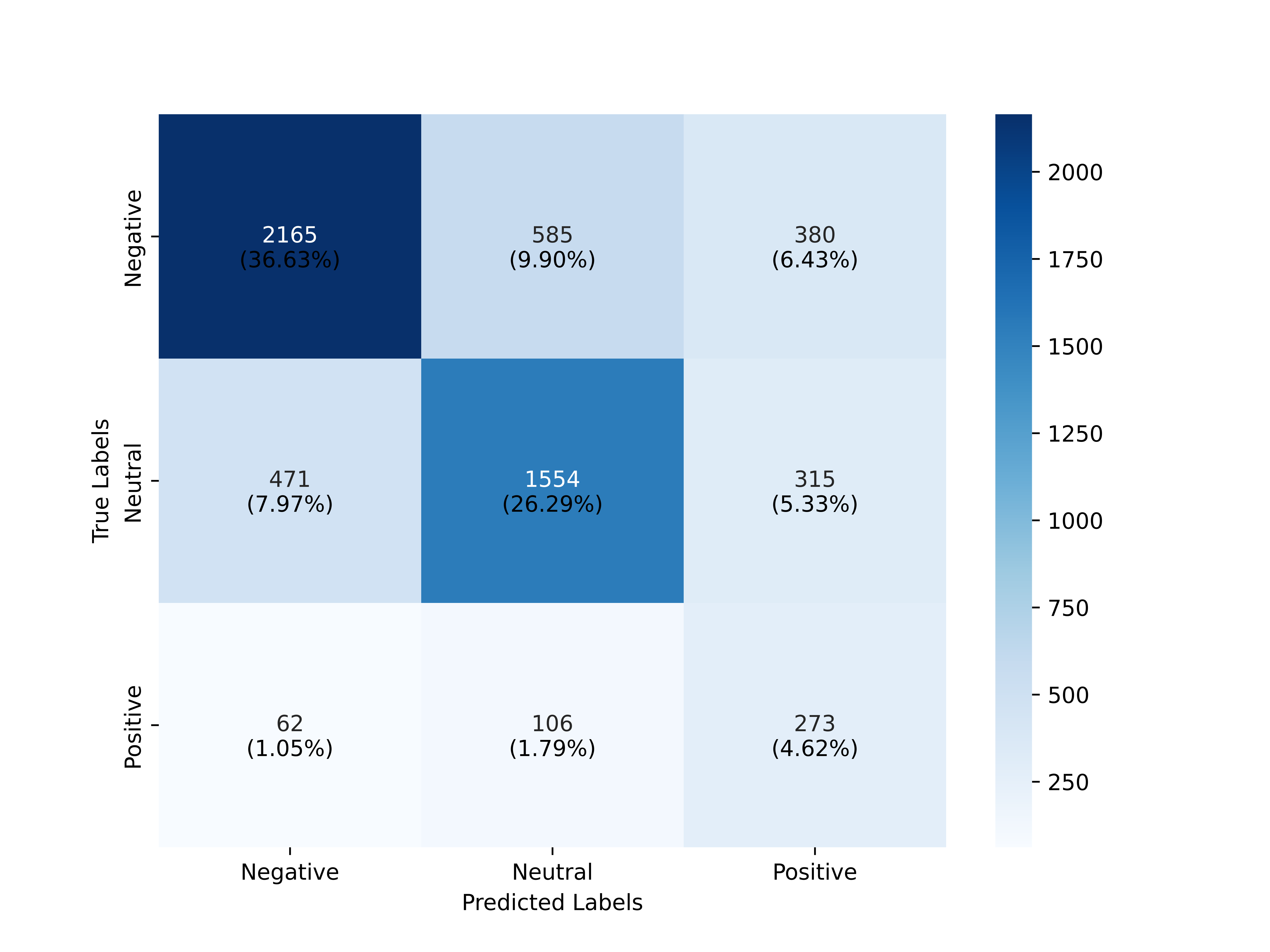}
    \caption{Falcon model evaluated on Augmented Data with reduced sarcasm.}
    \label{fig:falcon_hsa_aug}
  \end{subfigure}
  \caption{Confusion Matrices of the Falcon model finetuned on Nuclear Power data.}
  \label{fig:falcon_combined}
\end{figure}

These results suggest that LLMs trained on the nuclear power dataset are more attuned to sarcasm and gain significant improvements from sarcasm augmentation. This highlights the critical role of diverse, sarcasm-aware training data in improving an LLM's ability to handle complex linguistic nuances. For instance, Figure \ref{fig:falcon_hsa} displays the confusion matrix for the Falcon model fine-tuned on the nuclear power dataset and evaluated on the unaugmented human-labeled data. Figure \ref{fig:falcon_hsa_aug} shows the model’s performance when predicting sentiment after the dataset was augmented to reduce sarcasm. The results demonstrate a 16\% increase in accuracy for detecting negative sentiment and a reduction in confusion with positive and neutral sentiments, which are the highest confusions, suggesting that sarcastic tweets frequently convey negative sentiment in the context of nuclear power on social media. Additionally, GPT-3.5 was effective at paraphrasing sarcastic content, which enables the Falcon model to better classify sentiment accurately.

Finally, Table \ref{tab:sarcasm_ta} presents the performance of LLMs fine-tuned on data augmented using the TextAttack library for text augmentation \cite{morris2020textattack}, as outlined in Case 3 of Section \ref{sec:sarcasm_effect}. As expected, these models achieved the highest accuracy (significantly higher than Tables \ref{tab:sarcasm_np} and \ref{tab:sarcasm_gt}) when evaluated on the human-labeled test set, primarily because the synthetic data generated by TextAttack retains the same tone, content, and human labeling theme as the original human-labeled data. Furthermore, the LLM models showed minimal improvement from sarcasm augmentation, as indicated in row 2 of Table \ref{tab:sarcasm_ta}, with around 3\% increase in accuracy after augmentation. This is further supported by row 3 of Table \ref{tab:sarcasm_ta}, which shows that these models already performed well in predicting the sentiment of sarcastic tweets. Overall, these results suggest that adversarial text augmentation enhances both model robustness against sarcasm, significantly improves accuracy, and can be used instead of/reduce the need for labeling new tweets tediously by humans to be used for fine-tuning.

\begin{table}[!h]
\centering
\caption{LLM accuracy metrics when trained on the \textbf{TextAttack Augmented Dataset}. Model is evaluated on the test set of the Human Labelled Data (row 1), Human Labelled Data augmented by GPT-based text to reduce sarcasm (row 2), and Sarcastic Tweets only from Human Labelled Data (row 3).}
\begin{tabular}{lccc}
\toprule
\textbf{Evaluation Data}& \textbf{Falcon (\%)} & \textbf{Llama (\%)} & \textbf{Mistral (\%)} \\
\midrule
Human Labelled Data Without Sarcasm Augmentation& 81.5& 81.9& 76.7\\
Human Labelled Data With Sarcasm Augmentation& 84.2&   83.1& 79.32\\
Human Labelled Data Sarcastic Tweets Only & 84.4& 86.3& 83.8\\
\bottomrule
\end{tabular}
\label{tab:sarcasm_ta}
\end{table}

\subsection{Emoji sentiment analysis}

Table \ref{tab:emoji_comparison} presents the results of fine-tuning BERT-based models on datasets with and without emoji translation. Overall, most models show only slight improvements, with the highest gain of 2\% observed for the albert-base-v2 model. A similar pattern is seen with the bigger LLMs, as shown in Table \ref{tab:emoji_decoded_comparison}. While Falcon and Mistral achieve comparable accuracy in both scenarios, the LLaMA model shows a 3\% improvement when using emoji-translated data. Compared to LLM models, BERT variants had very low sentiment accuracy. The relatively small increase in accuracy for both BERT models and other LLMs suggests that incorporating emoji translations has limited impact on sentiment analysis for the nuclear power dataset—a rather unexpected finding. For LLMs, one hypothesis is that this could be due to their extensive pretraining on diverse and large-scale web data, which enhances their ability to handle varied text formats and contexts, thereby diminishing the added value of explicit emoji translations.

\begin{table}[!h]
\centering
\caption{Performance comparison of BERT-based models with and without emoji support based on the nuclear power tweet dataset.}
\begin{tabular}{lcc}
\toprule
\textbf{Model} & \textbf{Original (\%)} & \textbf{Emoji Decoded (\%)} \\
\midrule
albert-base-v2 & 45.0& 47.4\\
DeBERTa        & 45.9& 47.8\\
BERT           & 45.4& 47.0\\
\bottomrule
\end{tabular}
\label{tab:emoji_comparison}
\end{table}

\begin{table}[!htbp]
\centering
\caption{Performance comparison of LLM models with and without emoji support based on the nuclear power tweet dataset.}
\begin{tabular}{lcc}
\toprule
\textbf{Model} & \textbf{Original (\%)}& \textbf{Emoji Decoded (\%)} \\
\midrule
Falcon & 81.0& 81.8\\
Llama2 & 78.8& 81.7 \\
Mistral & 77.2& 76.3\\
\bottomrule
\end{tabular}
\label{tab:emoji_decoded_comparison}
\end{table}

In a similar manner to BERT models, the Llama-2, Falcon, and MistralAI models were fine-tuned on a subset of the nuclear power dataset, consisting of 77,439 tweets, all of which included emojis. Within this dataset, 55.4\% of the tweets expressed negative sentiment, 32.4\% positive sentiment, and 12.2\% neutral sentiment. Given this class imbalance, it is essential to evaluate additional metrics—namely, precision, recall, and F1-score—to accurately assess performance across all sentiment categories. These metrics for the Llama-2 model, fine-tuned on the emoji-translated version of the dataset, are presented in Table \ref{tab:llama_metrics}. The model performed well on both negative and positive sentiment labels. Since emojis are more frequently associated with strong emotional expressions rather than neutrality, this highlights the capability of LLMs to effectively interpret emoji-conveyed sentiment for these two categories.

\begin{table}[h!]
\centering
\caption{Llama2 performance for each sentiment label when fine-tuned using the nuclear power dataset with the emojis decoded.}
\begin{tabular}{lccc}
\toprule
\textbf{Label} & \textbf{Precision (\%)} & \textbf{Recall (\%)} & \textbf{F1-Score (\%)} \\
\midrule
Negative & 92.1& 87.4& 89.7\\
Neutral  & 41.4& 64.2& 50.3\\
Positive & 88.8& 78.4& 83.3\\
\bottomrule
\end{tabular}
\label{tab:llama_metrics}
\end{table}

\section{Discussions and Concluding Remarks}
\label{sec:conc}

This study explores the impact of various textual nuances, such as emojis and sarcasm, on sentiment analysis. To enhance the quality of the data, text paraphrasing techniques were also examined as a potential method to improve sentiment analysis results. In response to the lack of labeled data for sarcasm analysis, the authors created a human-labeled dataset as a ground truth, which was used to evaluate the performance of LLMs in different sarcasm scenarios. Several valuable insights and findings emerged from this study.

The results presented in Table \ref{tab:sarcasm_np} indicate that the nuclear power dataset, introduced by \cite{kwon2024sentiment}, demonstrates that topic-specific datasets are more vulnerable to sarcasm in sentiment analysis. External intervention, such as sarcasm removal, is necessary to improve model performance, which is clearly reflected in the \textbf{17\%} increase in sentiment accuracy for Falcon after sarcasm was removed from the tweets. The nuclear LLM models also struggled with accurately predicting the sentiment of sarcastic tweets, with accuracy hovering around 30\%.

In contrast, Table \ref{tab:sarcasm_gt} shows that when LLMs were trained on the general tweet dataset and tasked with predicting sentiment on the same data as the nuclear LLM models, the improvement in performance was relatively modest, ranging from \textbf{4\%} to \textbf{7\%}. Overall, this study highlights that for more effective sarcasm manipulation in domain-specific datasets, combining them with general tweet/text data can enhance their ability to recognize sarcasm and improve performance. General LLM models were found to be twice as accurate in predicting the sentiment of sarcastic tweets compared to the nuclear LLM models as Tables \ref{tab:sarcasm_np}-\ref{tab:sarcasm_gt} indicate.

For a more in-depth explanation of why LLMs fine-tuned on the general tweets outperformed in sarcasm detection the models fine-tuned on nuclear tweets, we examined the two datasets used for fine-tuning. Precisely, we used a fine-tuned LLM for sarcasm detection to count the number of sarcastic examples in both datasets. We used a BERT model available on huggingface, which was fine-tuned for binary classification to classify English text as sarcastic and non-sarcastic \cite{bertsar}. The model has a very good classification accuracy of 92\%, reported in the model card. The number of general tweets used for fine-tuning is around 532k; BERT classified around 35\% of them as sarcastic. In contrast, the number of nuclear energy tweets used for fine-tuning is around 484k; only 4\% of them are classified as sarcastic by BERT. As a result, this significant difference between the number of sarcastic tweets in the general and the nuclear datasets explains why the models fine-tuned on the general dataset outperformed those fine-tuned on the nuclear energy dataset.

The results from the TextAttack augmentation in Table \ref{tab:sarcasm_ta} revealed an interesting insight into fine-tuning LLMs. To enhance model accuracy and robustness against sarcasm, there is no need to collect additional data; instead, adversarial augmentation techniques can be employed to generate various versions of the same text by altering about 10\% of it while maintaining the overall content and sentiment. This adversarial text process is computationally efficient and can produce large amounts of synthetic data. One of the key findings of this study is that relying on binary classifiers to detect sarcasm, as many studies do, may be less effective and data-intensive compared to enriching the current dataset by generating synthetic variants.

\hl{As previously mentioned in Section \ref{sec:hld}, the original human-labeled dataset consisted of 10,000 tweets. However, tweets where the human annotators disagreed on the sentiment label were removed, resulting in a reduced dataset of 5,929 tweets. This implies an accuracy rate of approximately 59\% among human labelers. In comparison, LLMs fine-tuned on this dataset showed an accuracy 52\%—as illustrated in Table \ref{tab:sarcasm_np} for Llama-2, for example. This suggests that LLMs performed slightly worse than human annotators when classifying sentiment in the presence of sarcasm. While this similarity in performance might be coincidental rather than a definitive conclusion, it is worth noting that even human annotators did not achieve high agreement, reinforcing the notion that this sentiment labeling task is inherently difficult.}

Text paraphrasing successfully transformed approximately 40\% of low-confidence tweets (those with significant disagreement among labeling libraries) into high-confidence tweets (with increased agreement), demonstrating that paraphrasing can substantially enhance data quality and, in turn, improve sentiment analysis accuracy. This process further boosted LLM performance by 3-6\%, making the models more reliable. This study highlights the importance of incorporating paraphrasing as an additional text preprocessing step for sentiment analysis on social media data, where fragmented language and concise expressions are commonly used.

In the case of the dataset translated with emojis, the performance of LLMs fine-tuned on nuclear power tweets was observed to remain largely unchanged, indicating that the models did not leverage the emojis effectively. This finding contrasts with some other studies, but it highlights three key points within the context of nuclear power discussions on social media: first, the sentiment in the text may already be recognized by the LLM without the need for emojis, which may merely reinforce the sentiment. Second, it is unlikely that users will employ ambiguous language when discussing sensitive topics such as nuclear power or nuclear weapons, relying solely on emojis to convey their opinions. Third, since the emoji is replaced by up to 5 words only, as shown in Table \ref{table:emoji_text_mapping}, the attention mechanism in LLMs may not be affected by adding a few words to the text, especially if the text has a lot of words where the attention of LLMs will be distributed among them. Furthermore, the attention head of LLMs may find the added text to express emojis unrelated to the original text with emoji, and that is why it does not change its attention to these words, and hence the sentiment will not change. Thus, we can conclude that for our specific nuclear power dataset, which encompasses a wide range of tweets and nuclear sub-topics, emojis do not provide additional value for sentiment analysis.

In conclusion, this study emphasizes the importance of high-quality training data in both LLM performance and sentiment analysis outcomes. For social media data, strategies such as incorporating data diversity (rather than focusing on a single topic), using text paraphrasing to eliminate fragmented language, and applying sarcasm augmentation through adversarial text modifications or paraphrasing to remove sarcasm can significantly enhance performance. While decoding emojis may not negatively impact performance, it could also result in marginal improvements.

In our future research, we plan to integrate these language nuances into a real-time sentiment analysis dashboard to monitor public sentiment regarding sustainable and clean energy topics in the United States, including nuclear power and renewable energy, across various social media platforms such as Meta Threads, Reddit, Mastodon, and others.  \hl{We are also looking to design automated pipeline where input text data of low quality is paraphrased first before finetuning the LLM models to handle the social media text more efficiently.} Additionally, we aim to leverage the human-labeled dataset to explore the performance of LLMs in broader tasks like topic modeling, extending beyond the specific case studies addressed in this paper.

\section*{Data Availability}
\label{sec:data_avail}

The authors have all the data and codes to reproduce all the results in this work currently in a private GitHub repository. To ensure the confidentiality of this research, the authors will make this repository public during an advanced stage of the review process, which will be listed under our research group's public Github page: \url{https://github.com/aims-umich}

\section*{Acknowledgment}

This work is sponsored by the Department of Energy Office of Nuclear Energy under project number (DE-NE0009382), which is funded through the Nuclear Energy University Program (NEUP). This research also made use of Idaho National Laboratory computing resources, which are supported by the Office of Nuclear Energy of the U.S. Department of Energy under Contract No. DE-AC07-05ID14517.

\bibliographystyle{elsarticle-num}
\setlength{\bibsep}{0pt plus 0.3ex}
{
\bibliography{references}}
\end{document}